\documentclass[a4paper]{esannV2}
\usepackage{graphicx}
\usepackage[latin1]{inputenc}
\usepackage{amssymb,amsmath,array}
\usepackage{subcaption}
\usepackage{color}

\begin{document}
%style file for ESANN manuscripts
\title{A Sub-Layered Hierarchical Pyramidal Neural Architecture for Facial Expression Recognition}
%
%***********************************************************************
% AUTHORS INFORMATION AREA
%***********************************************************************
\author{Henrique Siqueira, Pablo Barros, Sven Magg, \\
Cornelius Weber and Stefan Wermter
%
% Optional short acknowledgment: remove next line if non-needed
\thanks{This work has received funding from the European Union under the SOCRATES project (No. 721619), and the German Research Foundation under the CML project (TRR 169).}
%
% DO NOT MODIFY THE FOLLOWING '\vspace' ARGUMENT
\vspace{.3cm}\\
%
% Addresses and institutions (remove "1- " in case of a single institution)
Knowledge Technology, Department of Informatics, University of Hamburg, \\
Vogt-Koelln-Str. 30, 22527 Hamburg, Germany
\\\\\large{\textbf{Pre-print version of \cite{siqueira2018sub}}}
}
%***********************************************************************
% END OF AUTHORS INFORMATION AREA
%***********************************************************************

\maketitle

\begin{abstract}
In domains where computational resources and labeled data are limited, such as in robotics, deep networks with millions of weights might not be the optimal solution. In this paper, we introduce a connectivity scheme for pyramidal architectures to increase their capacity for learning features. Experiments on facial expression recognition of unseen people demonstrate that our approach is a potential candidate for applications with restricted resources, due to good generalization performance and low computational cost. We show that our approach generalizes as well as convolutional architectures in this task but uses fewer trainable parameters and is more robust for low-resolution faces.
\end{abstract}

\section{Introduction}
% Facial expressions
For decades, several researchers from different areas have demonstrated interest in studying facial expressions, mainly, due to the strong relationship with human emotional states and intentions \cite{Tian2005}. Therefore, automatic emotion recognition approaches based on facial expressions have proven to be an essential part of complex cognitive architectures, such as social affective robots \cite{Rodrigues2015}.

% Recent Approaches for facial expressions recognition
Recently, deep networks have been successfully applied for facial expression recognition \cite{Khorrami2015}. Their capability of implicitly learning features from raw data avoids hand-engineering, and helps the model to learn emotional facial features from each person. The success of deep learning goes beyond facial expression, but most of the proposed architectures comprise millions of trainable parameters, demanding high computational power and a lot of labeled data for training, which may limit their application in domains where these resources are restricted.

% Pyramidal architectures for systems with limited resources
Pyramidal architectures are potential candidates for applications with limited resources, due to their good generalization performance and due to their low computational cost \cite{Phung2007, Fernandes2013}. The latter derives from their connectivity scheme where the weights are associated with the input neurons only, and from their hierarchical dimensionality reduction. However, as only one weight connects each input neuron to the next layer, the variety of input patterns that can be detected by each layer is limited, thereby compromising representational capability.

% Novelty
In this paper, we propose a novel connectivity scheme for a Hierarchical Pyramidal Neural Network (HPNN) that enhances generalization by increasing its capacity of learning features. The proposed connectivity allows building deeper pyramidal architectures by increasing the number of representations using the concept of sub-layers in each pyramidal layer. Otherwise, using the original connectivity scheme, additional pyramidal layers would lead to a reduction of the number of connections per layer to a point where the representational capability is drastically penalized. Finally, we evaluate the proposed approach regarding the number of trainable parameters and training samples for the task of learning facial expressions and generalizing for unseen people.

\section{Sub-Layered Hierarchical Pyramidal Neural Network}
% Introduction to pyramidal architectures
Pyramidal architectures are composed of two types of layers: pyramidal layers, which are responsible for developing hierarchical representations from raw images into a smaller spatial dimension, and responsible for the name of the architecture due to the shape of pyramid created by stacking pyramidal layers; and dense layers, which perform the classification based on these representations.

\begin{figure}[h]
	\centering
	\includegraphics[scale=0.45]{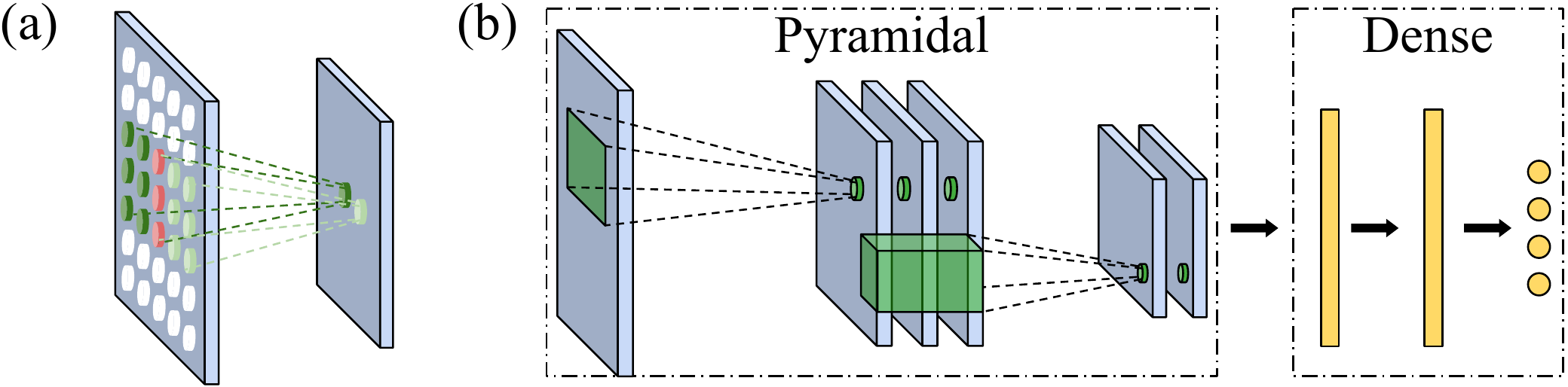}
	\caption{Illustration of the connectivity between pyramidal layers in (a), and the HPNN with its novel connectivity scheme to support sub-layers in (b).}
	\label{fig:hpnn}
\end{figure}

% Original architecture
In the original architecture by Phung et al. \cite{Phung2007}, the PyraNet, neurons in pyramidal layers are arranged in a matrix. Each neuron is connected to a local region in the previous layer, called receptive field. Adjacent neurons can be connected to an overlapping region in their receptive fields. An example of this connectivity is shown in Figure \ref{fig:hpnn}a. The green neurons on the right are connected to the projected 3 by 3 green regions on the left while the red neurons represent the overlapping region of one column shared between them. Each input neuron has only one weight associated to it. As a result, the weight matrix between two pyramidal layers $l-1$ and $l$ has the same shape as the layer $l-1$.

% Discussion about the connectivity scheme and problem
With this connectivity scheme, adjacent neurons are constrained to develop correlated spatial features from the previous layer, which is a desirable property for image classification since it is expected that neighboring regions might contain connected features. Contrary to convolutional layers, pyramidal layers have neurons with fixed receptive field positions, making their use appropriate for inputs where features should be found roughly in the same location, such as facial analysis. Nevertheless, the diversity of features that can be detected in each receptive field is limited since only one weight is connected to each input neuron. Consequently, the generality of the network might be limited in complex problems where different features may be presented to the receptive field.

% Proposed approach and Forward phase
Aiming to increase the diversity of features that can be detected in each input region, we propose a novel connectivity scheme for HPNNs to support a larger number of neurons connected to such a region by incorporating the concept of sub-layers in each pyramidal layer. As an example, the HPNN presented in Figure \ref{fig:hpnn}b comprises three pyramidal layers with one, three and two sub-layers each. Similarly to the filters in convolutional layers, each sub-layer in a given pyramidal layer has its own weight matrix, which yields sub-layers to learn different representations. The activation of a given neuron $y_{u, v}^{l, s}$ located at $(u, v)$ in a sub-layer $s$ in the pyramidal layer $l$ can be now computed as follows:
\begin{equation*}
y_{u, v}^{l, s} = f \left( \sum_{k=k_{min}}^{k_{max}} \sum_{i=i_{min}(u)}^{i_{max}(u)} \sum_{j=j_{min}(v)}^{j_{max}(v)} w_{k, i, j}^{l, s} y_{i, j}^{l-1, k} \right),
\label{eq:pyra_actv}
\end{equation*}
where $y_{i, j}^{l-1, k}$ is the input neuron connected to the neuron $y_{u, v}^{l, s}$ limited by the summation indexes $k$, $i$ and $j$, such that $k$ represents the index of a sub-layer in the layer $l-1$, and $i$ and $j$ represent the width and height of the receptive field.

% Backward phase: training details
To update the weight matrices between pyramidal layers, firstly, the error sensitivity $\delta_{i,j}^{l, k}$ of a given neuron $y_{i,j}^{l, k}$ can be obtained by:
\begin{equation*}
\delta_{i,j}^{l, k} = \sum_{s=s_{min}}^{s_{max}} \left( f'(x_{i, j}^{l, k}) w_{k, i, j}^{l+1, s} \sum_{u=u_{min}}^{u_{max}} \sum_{v=v_{min}}^{v_{max}} \delta_{u,v}^{l+1, s} \right),
\label{eq:pyra_sens}
\end{equation*}
where $f'(x_{i, j}^{l, k})$ is the derivative with respect to the input, $\delta^{l+1, s}$ is the error sensitivity of the neurons in the next pyramidal layer $(l+1, s)$ connected to the neuron $y_{i,j}^{l, k}$. Subsequently, the gradient of the weight $w_{k, i, j}^{l, s}$ that connects the neuron $y_{i, j}^{l-1, k}$ to the neuron $y_{u, v}^{l, s}$ in the next layer can be computed as given:
\begin{equation*}
\frac{\partial E}{\partial w_{k, i, j}^{l, s}} = y_{i, j}^{l-1, k} \sum_{u=u_{min}}^{u_{max}} \sum_{v=v_{min}}^{v_{max}} \delta_{u, v}^{{l, s}},
\label{eq:pyra_grad}
\end{equation*}
where $\delta_{u, v}^{{l, s}}$ is the error sensitivity of each neuron that contains the neuron $y_{i, j}^{l-1, k}$ in its receptive field. For simplification, the bias term has been omitted.

\begin{figure}[b]
	\centering
	\includegraphics[scale=0.45]{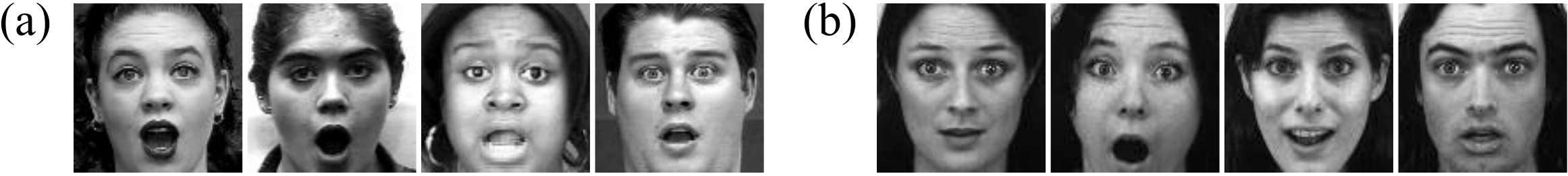}
	\caption{Surprise examples from the CK+ (a) and KDEF (b) datasets.}
	\label{fig:datasets}
\end{figure}

\section{Experiments and Results}
% Emotion recognition and the Extended Cohn-Kanade dataset
In our experiments, all networks are trained on the Extended Cohn-Kanade dataset (CK+, Figure \ref{fig:datasets}a) \cite{Lucey2010}. CK+ is composed of sequences of images, starting with a neutral face, and ending with one of seven universal facial expressions \cite{Ekman1989}: anger, contempt, disgust, fear, happiness, sadness and surprise.

% Base methodology
We evaluate using the subject-independent 10-fold cross-validation \cite{Khorrami2015}. With this methodology, we assess the generalization to unseen subjects, which more reliably indicates how the network would work in a real-world application than if the validation fold contained the same subjects as the training folds. The folds are populated by sampling subjects with a step size of ten after sorting them in ascending order based on their IDs. The first frame is labeled as neutral, whereas the last three frames are labeled as one of seven facial expressions, which results in eight categories. In each trial, one fold is used for test, one for validation, and eight for training. The images have been re-scaled to 96 by 96 pixels.

% First experiment: baseline
Firstly, we conduct a baseline experiment by comparing our approach with the original one by Phung et al. \cite{Phung2007}, and the convolutional architecture reported by Khorrami et al. \cite{Khorrami2015}, due to their good results on CK+, which comprises three convolutional layers with 64, 128, 256 filters, each of which is followed by a max-pooling layer, and on top, two dense layers with 300 and 8 neurons.

% Second experiment: generalization on the KDEF dataset
Secondly, we evaluate the generalization of the networks trained on CK+ for a different dataset of facial expressions since studies show that humans perceive and express emotions in many different ways, depending on their personal experience of feeling such an emotion \cite{Barrett2006}. The Karolinska Directed Emotional Faces (KDEF, Figure \ref{fig:datasets}b) \cite{Lundqvist1998} consists of images obtained in different angles from 70 subjects conveying the same emotional categories presented in CK+, except for the contempt category. The neutral category is also present, resulting in seven categories. Only the centralized face images from KDEF are used.

% Third experiment: reduced number of subjects for traning
Focusing on evaluating architectures for applications with restricted resources, in our third experiment, we investigate the generalization performance when the number of subjects for training is reduced by half. Therefore, in each trial of the 10-fold cross-validation, four of the eight folds reserved for training are removed. On average, the number of images for training are reduced from 1046.4 to 577.6.

Furthermore, we investigate the robustness for low-resolution facial expressions, which can be a result of a face detected at a long distance in real-world applications. The low-resolution images are generated by applying a mean filter varying its size from 3 to 15 with a step size of 3. In this fourth experiment, the networks trained on CK+ using eight folds with the original images are tested to the respective fold in each trial, but with blurred images.

\begin{figure}[h]
	\centering
	\includegraphics[scale=0.43]{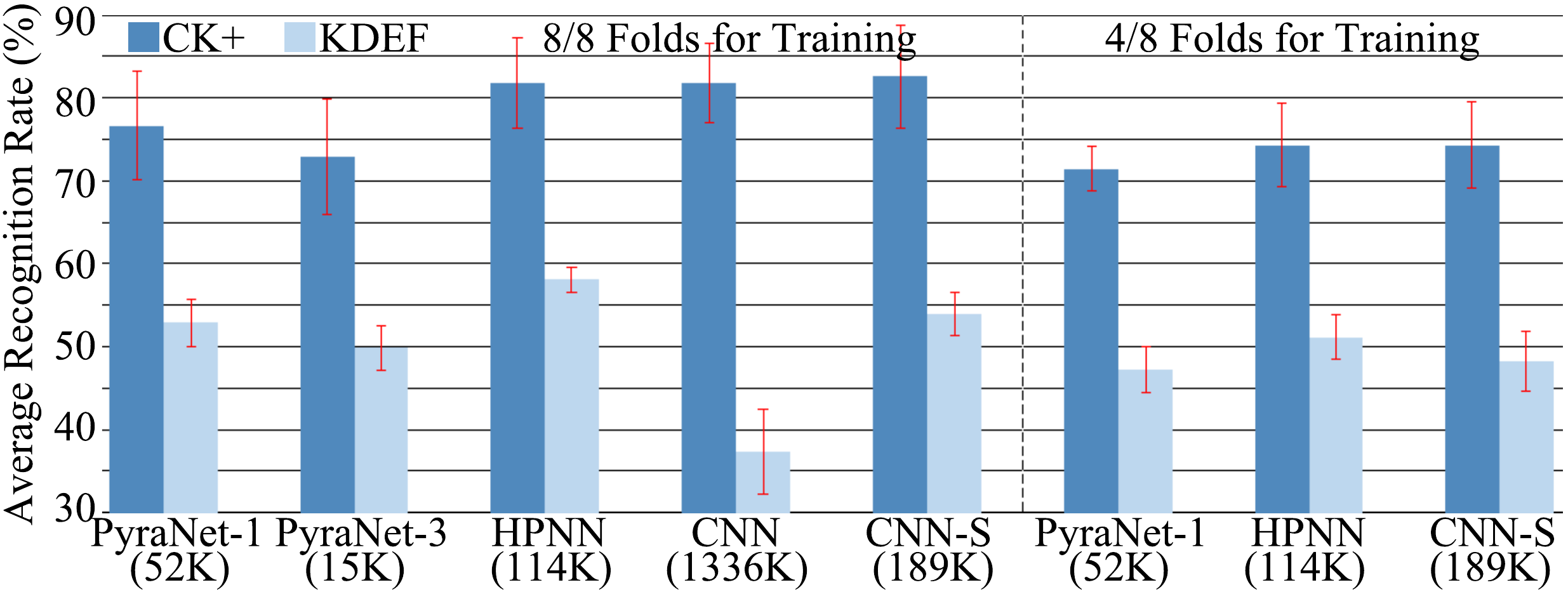}
	\caption{Comparison results on the CK+ and KDEF datasets with the number of trainable parameters used for each architecture.}
	\label{fig:folds}
	\vspace{-12pt}
\end{figure}

% First experiment: base architecture
Figure \ref{fig:folds} shows the average recognition rates for the first three experiments. Initially, different HPNN architectures were tested to reach a generalization performance equivalent to the standard CNN reported by Khorrami et al. \cite{Khorrami2015} (CNN in Figure \ref{fig:folds}). The architecture has three pyramidal layers with 4, 8 and 8 sub-layers, followed by two dense layers with 40 and 8 neurons. The HPNN has achieved the same recognition rate on CK+ as the CNN, i.e. 81.8\%, but contains far less trainable parameters: 113,520 compared to 1,335,956 of the CNN. The PyraNet-3 with three pyramidal layers, but with only one sub-layer, has achieved the lowest recognition rate of 72.90\%. If we remove the last two pyramidal layers of the PyraNet-3, the recognition rate increases about 4\% as shown in Figure \ref{fig:folds} by the PyraNet-1. This result shows the problem of going deeper with pyramidal architectures using the original approach by Phung et al. \cite{Phung2007}.

% Second experiment: Advantage on KDEF
The recognition rates decreased when the networks trained on CK+ were tested on KDEF, as expected. In this case, the HPNN has reached the highest recognition rate and the lowest standard deviation of 58.07\% (1.55\%), which results in 20.76\% more samples correctly classified than the CNN. The excessive number of trainable parameters in the CNN might have caused overfitting on CK+, and consequently, poor generalization to KDEF. Therefore, we gradually reduced the number of trainable parameters until the point before the CNN performance starts to degenerating on CK+. This smaller architecture is referred in our experiments as the CNN-S, and has three convolutional layers with 24, 12, 8 filters, each of which is followed by a max-pooling layer, and on top, two dense layers with 40 and 8 neurons. The recognition rate increased to 82.6\% for the test set on CK+, and increased to 53.99\% (2.62\%) on the KDEF dataset.

% Third experiment: Reduced number of subjects
When the networks are trained with half of the subjects on CK+, the recognition rates on the test set decreased 7\% for both the HPNN and the CNN. As in the previous experiment, the HPNN achieved the highest recognition rate on the KDEF dataset with 51.15\% (2.67\%), followed by the CNN-S with 48.22\% (3.65\%), and the PyraNet-1 with 47.25\% (2.75\%).

\begin{figure}[h]	
	\centering
	\includegraphics[scale=0.54]{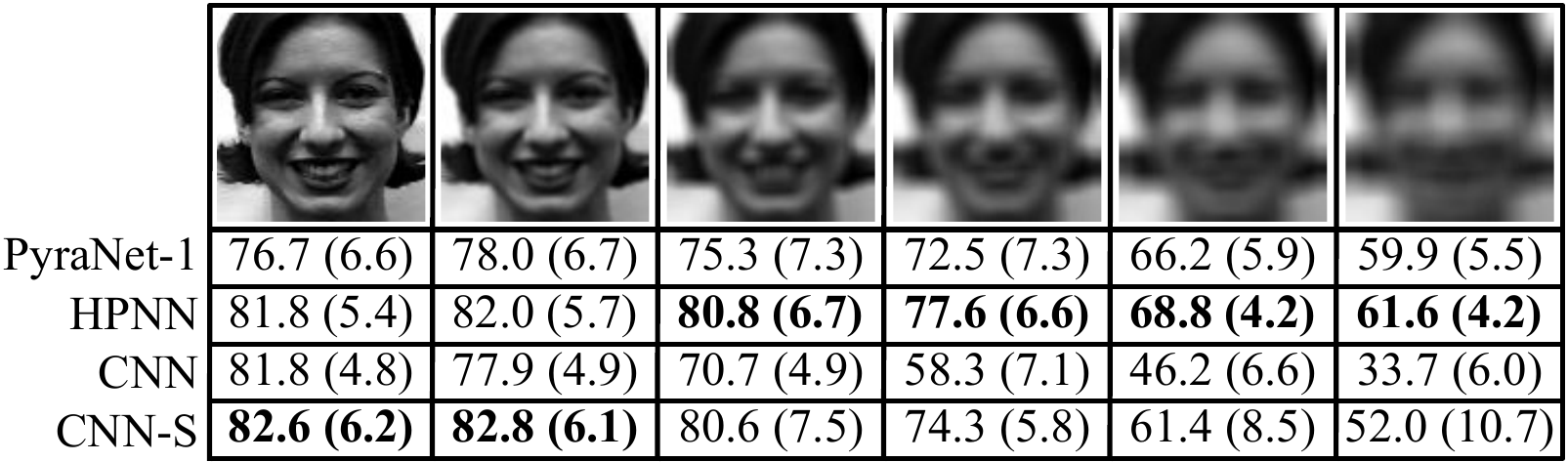}
	\caption{Average recognition rates (\%) and the standard deviation in parentheses over increasing degrees of blur.}
	\label{fig:smooth}
\end{figure}

% Fourth experiment
Finally, Figure \ref{fig:smooth} presents the average recognition rates and the standard deviation in parentheses for the test set with low-resolution facial expressions of the CK+ dataset. Note that the CNN with millions of trainable parameters most quickly loses its capability of classifying facial expressions.  The HPNN is the most robust architecture for the classification of low-resolution facial expressions, correctly classifying 61.63\% samples in the most challenging scenario.

\section{Conclusions and Future Works}
% Prior knowledge
How can we build neural architectures for applications with restricted resources? One strategy is to consider some prior knowledge about the task during the design phase. For pyramidal architectures, the size of receptive fields connected to the input image should be large enough to detect relevant facial features. This increases the efficiency of pyramidal networks by decreasing the number of receptive fields connected to small regions without any relevant information for the task, such as skin, hair and background.

% Conclusions
In this paper, we proposed a novel connectivity scheme for Hierarchical Pyramidal Neural Networks, which has improved their recognition rates by increasing the diversity of features that can be detected by each pyramidal layer. The HPNN has achieved similar generalization performance as convolutional architectures on facial expression recognition on CK+ but using fewer trainable parameters. Consequently, the network consumes less memory and decreases the likelihood of overfitting to a certain dataset. We demonstrate this effect by training the networks on CK+ and testing on a different dataset of facial expressions of emotion, the KDEF dataset, where the HPNN has shown higher performance.

% Why smooth images
Pyramidal architectures with fixed receptive fields preserve the spatial information of the features. Therefore, not only local features are detected but also the information about where they are located. This property provides a certain advantage for the HPNN over the CNN on the classification of low-resolution facial expressions. This finding motivates us, as future work, to evaluate the combination of convolutional and pyramidal layers to leverage specific and spatially related features in the same architecture.

% ****************************************************************************
% BIBLIOGRAPHY AREA
% ****************************************************************************
\begin{footnotesize}
\bibliographystyle{unsrt}
\bibliography{esannV2}
\end{footnotesize}

% ****************************************************************************
% END OF BIBLIOGRAPHY AREA
% ****************************************************************************
\end{document}